\documentclass[10pt,twocolumn,letterpaper]{article}
\usepackage{iccv}
\usepackage{times}
\usepackage{epsfig}
\usepackage{graphicx}
\usepackage{amsmath}
\usepackage{amssymb}
\usepackage{amsfonts}
\usepackage{latexsym}
\usepackage{mathrsfs}
\usepackage{booktabs}
\usepackage{algorithm}
\usepackage{algorithmic}
\usepackage{caption}
\usepackage{color}
\usepackage[noblocks]{authblk}
\usepackage{subfig}
\makeatletter
\renewcommand\normalsize{%
   \@setfontsize\normalsize\@xpt\@xiipt
   \abovedisplayskip 4\p@ \@plus2\p@ \@minus5\p@
   \abovedisplayshortskip \z@ \@plus3\p@
   \belowdisplayshortskip 5\p@ \@plus3\p@ \@minus1\p@
   \belowdisplayskip \abovedisplayskip
   \let\@listi\@listI}
\makeatother

\iccvfinalcopy 

\def\iccvPaperID{****} 

\ificcvfinal\fi
\begin{document}

\title{Integrating Boundary and Center Correlation Filters for Visual Tracking with Aspect Ratio Variation}

\author[1]{\large Feng Li}
\author[1]{\large Yingjie Yao}
\author[2]{\large Peihua Li}
\author[3]{\large David Zhang}
\author[,1]{\large Wangmeng Zuo \thanks{Corresponding author.}}
\author[4]{\large Ming-Hsuan Yang}

\affil[1]{\normalsize School of Computer Science and Technology, Harbin Institute of Technology, China}
\affil[2]{\normalsize School of Information and Communication Engineering, Dalian University of Technology, China}
\affil[3]{\normalsize Department of Computing, The Hong Kong Polytechnic University, China}
\affil[4]{\normalsize School of Engineering, University of California, Merced, USA \authorcr{\tt \small{fengli\_hit@hotmail.com, yyjhit@sina.com, peihuali@dlut.edu.cn

csdzhang@comp.polyu.edu.hk, wmzuo@hit.edu.cn, mhyang@ucmerced.edu}}}

\maketitle

\begin{abstract}
The aspect ratio variation frequently appears in visual tracking and has a severe influence on performance. Although many correlation filter (CF)-based trackers have also been suggested for scale adaptive tracking, few studies have been given to handle the aspect ratio variation for CF trackers. In this paper, we make the first attempt to address this issue by introducing a family of 1D boundary CFs to localize the left, right, top, and bottom boundaries in videos. This allows us cope with the aspect ratio variation flexibly during tracking. Specifically, we present a novel tracking model to integrate 1D Boundary and 2D Center CFs (IBCCF) where boundary and center filters are enforced by a near-orthogonality regularization term. To optimize our IBCCF model, we develop an alternating direction method of multipliers. Experiments on several datasets show that IBCCF can effectively handle aspect ratio variation, and achieves state-of-the-art performance in terms of accuracy and robustness. The source code of our tracker is available at {\url{https://github.com/lifeng9472/IBCCF/}}.
\end{abstract}

\section{Introduction}
\begin{figure}[!htbp]
\centering
\setlength{\abovecaptionskip}{0.2cm}
\setlength{\belowcaptionskip}{-0.5cm}
\subfloat{%
  \includegraphics[width=1\linewidth]{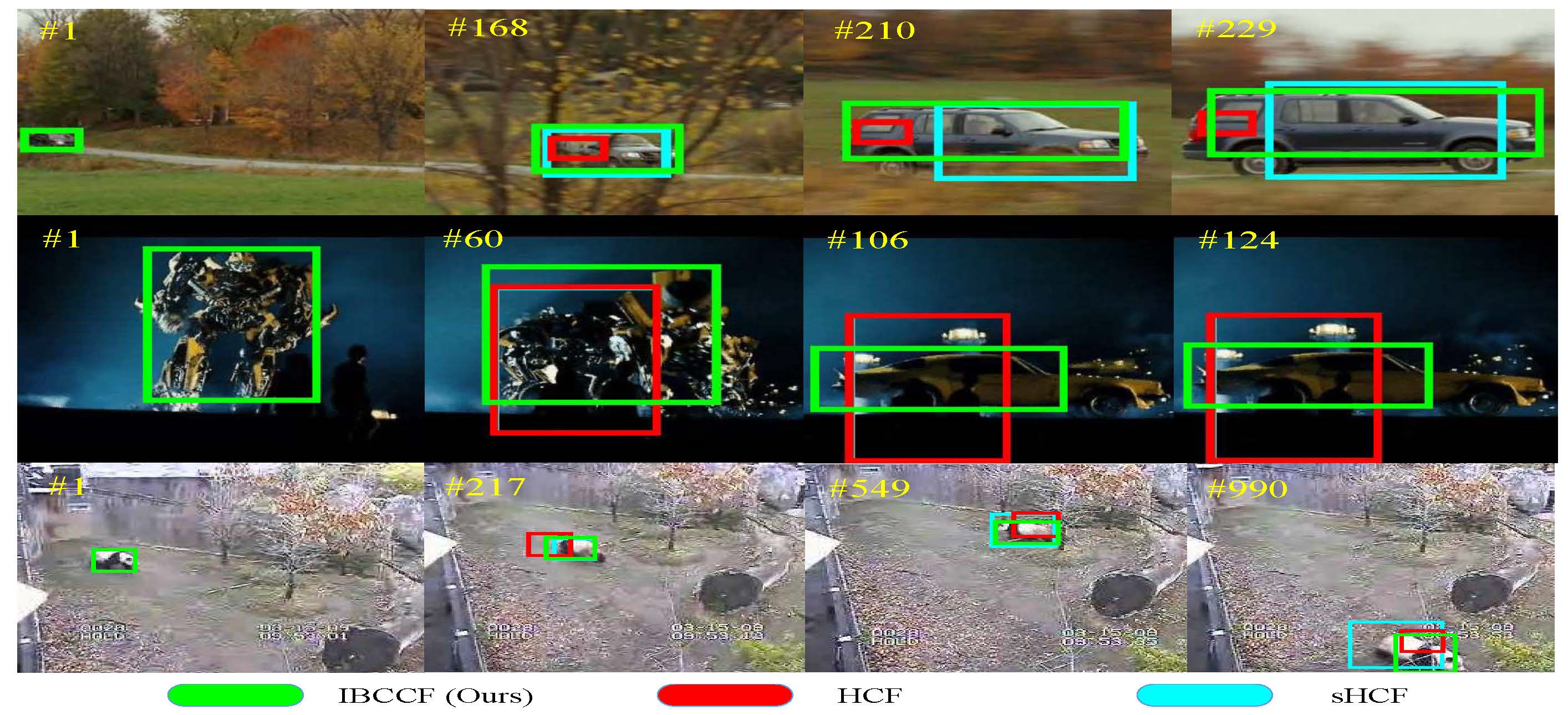}}\\
\caption{Three example sequences (\emph{CarScale}, \emph{Trans} and \emph{Panda}) with aspect ratios variation in OTB benchmark. Given the initial target position in the first column, we compare our method (IBCCF) with its counterparts HCF \cite{ma2015hierarchical} and sHCF. The results show our IBCCF is superior to HCF and sHCF in the case of aspect ratio variation.}
\label{fig:QualitativeResults}
\end{figure}
Visual tracking is one of the most fundamental problems in computer vision.
And it plays a critical role in versatile applications such as video surveillance, intelligent transportation, and human-computer interaction \cite{Yang2011Recent,Smeulders2014,wu2015object,Zhang2015Robust,vcehovin2016visual}.
Given its annotation on the initial frame, visual tracking aims to estimate the trajectory of a target with large appearance variations caused by many factors, \eg, scale, rotation, occlusion, and background clutter.
Although great advance has been made, it remains a challenging issue to develop an accurate and robust tracker to handle all these factors.
Recently, correlation filter (CF)-based approaches have received considerable attention in visual tracking \cite{bolme2010visual,henriques2015high,danelljan2016discriminative}.
In these methods, a discriminative correlation filter is trained to generate 2D Gaussian shaped responses centered at target position.
Benefited from the circulant matrix and fast Fourier transform (FFT), the CF-based trackers usually perform very efficiently.
With the introduction of deep features \cite{ma2015hierarchical,qi2016hedged} and spatial regularization \cite{danelljan2015learning,lukevzivc2016discriminative} and continuous convolution \cite{Danelljan2016CCOT}, the performance of CF-based trackers have been persistently improved, and lead to the state-of-the-art performance.

Despite the advances in CF-based tracking, the aspect ratio variation remains an open problem.
The changes in aspect ratio can be caused by variations in-plane/out-of-plane rotation, deformation, occlusion, or scale variation, and usually have a severe effect on tracking performance.
To handle scale variation, Li and Zhu \cite{li2014scale} propose a Scale Adaptive with Multiple Features tracker (SAMF).
Danelljan et al. \cite{danelljan2016discriminative} suggest a discriminative scale space tracking (DSST) method to learn separate CFs for translation and scale estimation on scale pyramid representation.
Besides, scale variation issue can also be handled by the part-based CF trackers \cite{Liu2015Real,Li2015RPT,Liu2016SCF}.
These methods, however, can only cope with scale variation issue but cannot well address the the aspect ratio variation issue.
Fig. \ref{fig:QualitativeResults} illustrates the tracking results on three sequences with aspect ratio variation.
It clearly shows that, even with deep CNN features, neither the standard CF (\ie HCF \cite{ma2015hierarchical}) tracker nor its scale-adaptive version (sHCF) can address the issue of aspect ratio variation caused by in-plane/out-of-plane rotation and deformation.

In this paper, we present a visual tracking model to handle aspect ratio variation by integrating boundary and center correlation filters (IBCCF).
The standard CF estimates the trajectory by finding the highest response in each frame to locate the center of the target, and can be seen as a center tracker.
In contrast, we introduce a family of 1D boundary CFs to localize the positions of the left, right, top and bottom boundaries (\ie $m_l, m_r, n_t, n_b$) in the sequences, respectively.
By treating 2D boundary region as a multi-channel representation of 1D vectors, a boundary CF is learned to generate 1D Gaussian shaped responses centered at target boundary (See Fig. \ref{fig:Comparisonof1Dand2D}).
By using boundary CFs, the left, right, top and bottom boundaries can be flexibly tuned in the image sequences.
Thus the aspect ratio variation can be naturally handled during tracking.

We empirically analyze and reveal the near-orthogonality property between the center and boundary CFs.
Then, by enforcing the orthogonality with an additional regularization term (\ie near-orthogonality constraint), we present a novel IBCCF tracking model to integrate 1D boundary and 2D center CFs.
Meanwhile, an alternating direction method of multipliers (ADMM) \cite{boyd2011distributed} is then developed to optimize the proposed IBCCF model.

To evaluate our IBCCF, extensive experiments have been conducted on OTB-2013, OTB-2015 \cite{wu2015object} , Temple-Color \cite{Liang2015Encoding}, VOT-2016 \cite{vcehovin2016visual} and VOT-2017 datasets.
The results validate the effectiveness of IBCCF on handling aspect ratio variation.
Compared with several state-of-the-art trackers, our IBCCF achieves comparable performance in terms of accuracy and robustness.
As shown in Fig. \ref{fig:QualitativeResults}, by using CNN features, our IBCCF can well adapt to the aspect ratio variation in the three
sequences, yielding better tracking performance.

To sum up, the contributions of this paper are three-fold:
\begin{itemize}
  \item A novel IBCCF model is developed to address the aspect ratio variation in CF-based trackers. To achieve this, we first introduce a family of boundary CFs to track the left, right, top and bottom boundaries besides tracking the target center.
  Then, we combines boundary and center CFs by encouraging orthogonality between them for accurate tracking.
  \item An ADMM algorithm is suggested to optimize our IBCCF model, where each subproblem has the closed-form solution. Our algorithm alternates between
  updating center CFs and updating boundary CFs, and empirically converges with very few iterations.
  \item The extensive experimental results demonstrate the effectiveness of our proposed IBCCF, and it achieves comparable tracking performance against several state-of-the-art trackers.
\end{itemize}
\section{Related Work}\label{relatedWork}

In this section, we provide a brief survey on CF-based trackers, and discuss several scale adaptive and part-based CF trackers close to our method.
\subsection{Correlation Filter Trackers}
Denote by $\mathbf{x}$ an image patch of $M \times N$ pixels.
Let $\mathbf{y}$ be a 2D Gaussian shaped labels.
The correlation filter $\mathbf{w}$ is then learned by minimizing the ridge regression objective:
\begin{align}\label{equ:CF}
\mathbf{w} \!=\! \arg \min_{\mathbf{w}} \left\{ \mathcal{E}(\mathbf{w}) \!=\! \left \|\mathbf{w} \otimes_2 \mathbf{x} \!-\! \mathbf{y}\right \|^2
\!+\! \lambda\left \| \mathbf{w} \right \|^2 \right\},
\end{align}

where $\lambda$ denotes the regularization parameter, and $\otimes_2$ is the 2D convolution operator.
Denote by $\hat{\mathbf{x}}$ the Fourier transform of ${\mathbf{x}}$, and $\hat{\mathbf{x}}^{*}$ the complex conjugate of $\hat{\mathbf{x}}$.
Using fast Fourier transform (FFT), the closed-form solution to Eqn. (\ref{equ:CF}) can be given as:
\begin{align}\label{equ:CF_solution}
\mathbf{w} = \mathcal{F}^{-1} \left(\frac{ \hat{\mathbf{x}}^{*} \odot \hat{\mathbf{y}} }{\hat{\mathbf{x}}^{*} \odot \hat{\mathbf{x}} + \lambda}\right),
\end{align}
where $\odot$ denotes the element-wise multiplication, and $\mathcal{F}^{-1} (\cdot)$ represents the inverse Discrete Fourier transform operator.

From the pioneering MOSSE by Bolme et al. \cite{bolme2010visual}, great advances have been made in CF-based tracking.
Henriques et al. \cite{henriques2015high} extend MOSSE to learn nonlinear CF via kernel trick.
And the multi-channel extension of CF has been studied in \cite{kiani2013multi}.
Driven by feature engineering, HOG \cite{Dalal2005Histograms}, color names \cite{Danelljan2014Adaptive} and deep CNN features \cite{ma2015hierarchical,qi2016hedged} have been successively adopted in CF-based tracking.
Other issues, such as long-term tracking \cite{Ma2015Long}, continuous convolution \cite{Danelljan2016CCOT}, spatial regularization \cite{danelljan2015learning,lukevzivc2016discriminative}, and boundary effect \cite{kiani2015correlation}, are also investigated to improve
tracking accuracy and robustness.
Besides ridge regression, other learning models, e.g., support vector machine (SVM) \cite{zuo2016learning,ningobject} and sparse coding \cite{sui2016real}, are also introduced.
Due to the page limits, in the following, we further review the scale adaptive and part-based CFs which are close to our work.

\subsection{Scale Adaptive and Part-based CF Trackers}

The first family of methods close to our approach are scale adaptive CF trackers, which aim to estimate target scale changes during tracking.
Among the current scale adaptive CF trackers \cite{danelljan2015learning,Danelljan2016CCOT,Ma2015Long}, SAMF \cite{li2014scale} and DSST \cite{danelljan2016discriminative} are two commonly used methods for scale estimation.
They apply the learned filter to samples of multi-resolutions around the target, and compute the response for each scale of the sample whose the maximum response is seen as the optimal scale.
However, such strategy is time consuming in the case of large scale space, and many improvements over them have been proposed. Tang and Feng \cite{Tang2015Multi} employ bisection search and fast feature scaling method to speed up scale space searching.
Bibi and Ghanem \cite{Bibi2015Multi} maximize the posterior probability rather than the likelihood (\ie maximum response map) in different scales for more stable detections.
Additionally, Zhang et al. \cite{Zhang2014} also suggest a robust scale estimation method by averaging the scales over $n$ consecutive frames.
Despite these successes in isometric scale variation, such kind of methods cannot well address aspect ratio variation. Different from the aforementioned methods, the proposed IBCCF approach can handle aspect ratio variation effectively with the introduction of boundary CFs.

Our IBCCF also shares some philosophy with the part-based CF methods, which divide the entire target into several parts and merge the results from all parts for final prediction.
For example, Liu et al. \cite{Liu2015Real} divide the target into five parts which are assigned with five independent CF trackers, and the final target position estimation is obtained by merging five CF trackers using Bayesian inference methods.
Different from simple dividing parts, Li et al. \cite{Li2015RPT} propose to exploit reliable parts, which estimate their probability distributions under a sequential Monte Carlo framework and employ a Hough voting scheme to locate the target.
In the similar line, Liu et al. \cite{Liu2016SCF} propose to jointly learn multiple parts from the target with CF trackers in an ADMM framework.
Compared with the part-based trackers, the proposed IBCCF has several merits: (1) IBCCF chooses to track meaningful boundary regions, which is more general than the fixed partition based method \cite{Liu2015Real} and easier to be handled than the learned parts based method \cite{Liu2016SCF};
(2) With the introduction of 1D boundary CFs, IBCCF can naturally deal with the aspect ratio variation problem;
(3) The near-orthogonality constraint between boundary and center CFs encourages IBCCF better performance than part-based ones.
\begin{figure}[htbp]
\centering
\setlength{\abovecaptionskip}{0.1cm}
\setlength{\belowcaptionskip}{-0.4cm}
\subfloat{%
  \includegraphics[width=1\linewidth]{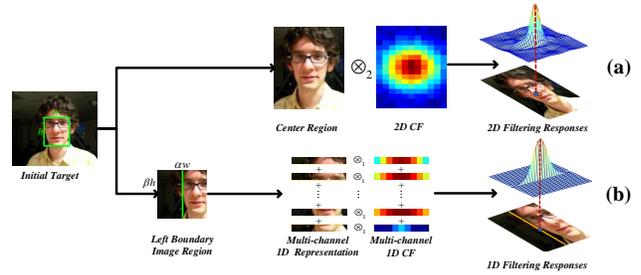}}\\
\caption{Comparison between standard CF and the proposed 1D boundary CF. Note that $\otimes_1$ and $\otimes_2$ represent the 1D and 2D convolution, respectively.
(a): Standard CF convolves the center region with a 2D CF to generate 2D responses for finding the center position.
(b): 1D boundary CF crops the boundary region centered at target boundary, then 2D boundary region
is reshaped as a multi-channel representation of 1D vectors and convolved with 1D CF to produce 1D responses for finding the boundary position.}
\label{fig:Comparisonof1Dand2D}
\end{figure}
\section{The Proposed IBCCF Model}\label{OurModel}

In this section, we first introduce the boundary correlation filters.
Then, we investigate the near-orthogonality property between the boundary and center CFs, and finally present our IBCCF model.

\subsection{Boundary Correlation Filters}

In standard CF, the bounding box of a target with fixed size is uniquely characterized by its center $(m_c, n_c)$.
By incorporating with scale estimation, the target bounding box can be determined by both the center $(m_c, n_c)$ and the scale factor $\alpha$.
However, both standard and scale adaptive CFs cannot address the aspect ratio variation issue, so better description of bounding box is required.
For CNN-based object detection \cite{Ross2014}, the bounding box is generally parameterized by center coordinate, its height and width.
Although such parameterization scheme can cope with aspect ratio variation, it is difficult to predict target height and width in the CF framework.

In this work, the bounding box is parameterized with its left, right, top and bottom boundaries $\mathcal{B} = \{m_l, m_r, n_t, n_b\}$.
It is natural to see that such parameterization is able to handle aspect ratio variation with dynamically adjusting four boundaries of target.
Moreover, for each boundary of $\mathcal{B}$, a 1D boundary CF (BCF) is learned to estimate the left, right, top or bottom boundary, respectively.
Taking the left boundary $m_l$ as an example, Fig. \ref{fig:Comparisonof1Dand2D}(b) illustrates the process of 1D boundary CF.
Given a target bounding box, let $(m_c, n_c)$ be the center, $h$ and $w$ be the height and width.
Its left boundary can be represented as $m_l = m_c-\frac{w}{2}$.
Then we crop a left boundary image region $\mathbf{x}_l$ centered at $(m_l, n_c)$ with width $w_l = \alpha w$ and height $h_l = \beta h$.

To learn 1D boundary CF, the left boundary image region is treated as a multi-channel (\ie $h_l$) representation of 1D $w_l \times 1$ vectors
$\mathbf{x}_l = [\mathbf{x}_l^1; ...; \mathbf{x}_l^{h_l}]$.
Denote by $\mathbf{y}_l$ a 1D Gaussian shaped labels centered at $m_l$.
Then the 1D left boundary CF model can then be formulated as,
\begin{align}\label{equ:BCFFomulation}
\mathbf{w}_{l} \!\!=\!\! \arg \min_{\mathbf{w}_{l}}  \mathcal{L}_l(\mathbf{w}_{l}) \!\!=\!\! \left \| \sum_{j=1}^{h_l} \mathbf{x}_l^j \!\otimes_1\! \mathbf{w}_l^j \!-\! \mathbf{y}_l \right \|^2
\!\!+\!\! \lambda \left \| \mathbf{w}_l \right \|^2
\end{align}
where $\otimes_1$ denotes the 1D convolution operator.
For each channel of $\mathbf{w}_{l}$, its closed form solution $\mathbf{w}_{l}^j$ can be obtained by,
\begin{align}\label{equ:BCF_solution}
\mathbf{w}_{l}^j = \mathcal{F}^{-1} \left(\frac{ \hat{\mathbf{x}}_{l}^{j*} \odot \hat{\mathbf{y}_l} }{\sum_{j=1}^{h_l} \hat{\mathbf{x}}_{l}^{j*} \odot \hat{\mathbf{x}}_{l}^{j*} + \lambda}\right).
\end{align}

As shown in Fig. \ref{fig:Comparisonof1Dand2D}(a), the center region is convolved with a 2D CF to generate a 2D filtering responses.
Then, the target center is determined by the position with the maximum response.
Thus standard CF can be seen as a center CF (CCF) tracker.
In contrast, as shown in Fig. \ref{fig:Comparisonof1Dand2D}(b), the left boundary region is first equivalently written as a multi-channel representation of 1D vectors.
The multi-channel 1D vectors are then convolved with multi-channel 1D correlation filters to produce 1D filtering responses.
And the left boundary is determined by finding the position with the maximum response.
Analogously, the other boundaries can also be obtained to track with the right, top and bottom boundary CFs, respectively. Fig. \ref{fig:CommmonRegion} shows the setting of the boundary regions based on the target bounding box.

When a new frame comes, we first crop the boundary regions, which are convolved with the corresponding boundary CFs.
The left, right, top and bottom boundaries are then determined based on the corresponding 1D filtering responses.
Note that each boundary is estimated independently.
Thus, our BCF approach can adaptively fit target scale and aspect ratio.
\subsection{Near-orthogonality between Boundary and Center CFs}
It is natural to see that the boundary and center CFs are complementary and can be integrated to boost tracking performance. To locate target in the current frame, we can first detect an initial position with CCF, then BCFs are employed to further refine the boundaries and position.
To update the tracker, we empirically investigate the relationship between the CCF and BCFs, and then suggest to include a near-orthogonality regularizer for better integration.

Suppose that the size of left boundary region for BCF is the same with that of center region for CCF.
Without loss of generality, we let $[\mathbf{x}]_{\text{vec}}$, $[\mathbf{w}]_{\text{vec}}$, and $[\mathbf{w}_l]_{\text{vec}}$ be the vectorization of the center region, center CF, and left boundary CF, respectively.
On one hand, the filtering responses of left boundary CF should be higher at the left boundary and near zero otherwise.
So we have that $[\mathbf{w}_l]_{\text{vec}}^T [\mathbf{x}]_{\text{vec}} \approx 0$, which indicates that $[\mathbf{w}_l]_{\text{vec}}$ and $[\mathbf{x}]_{\text{vec}}$ are nearly orthogonal.
\begin{figure}[htbp]
\centering
\setlength{\abovecaptionskip}{0.1cm}
\setlength{\belowcaptionskip}{-0.2cm}
\subfloat{%
  \includegraphics[width=0.6\linewidth]{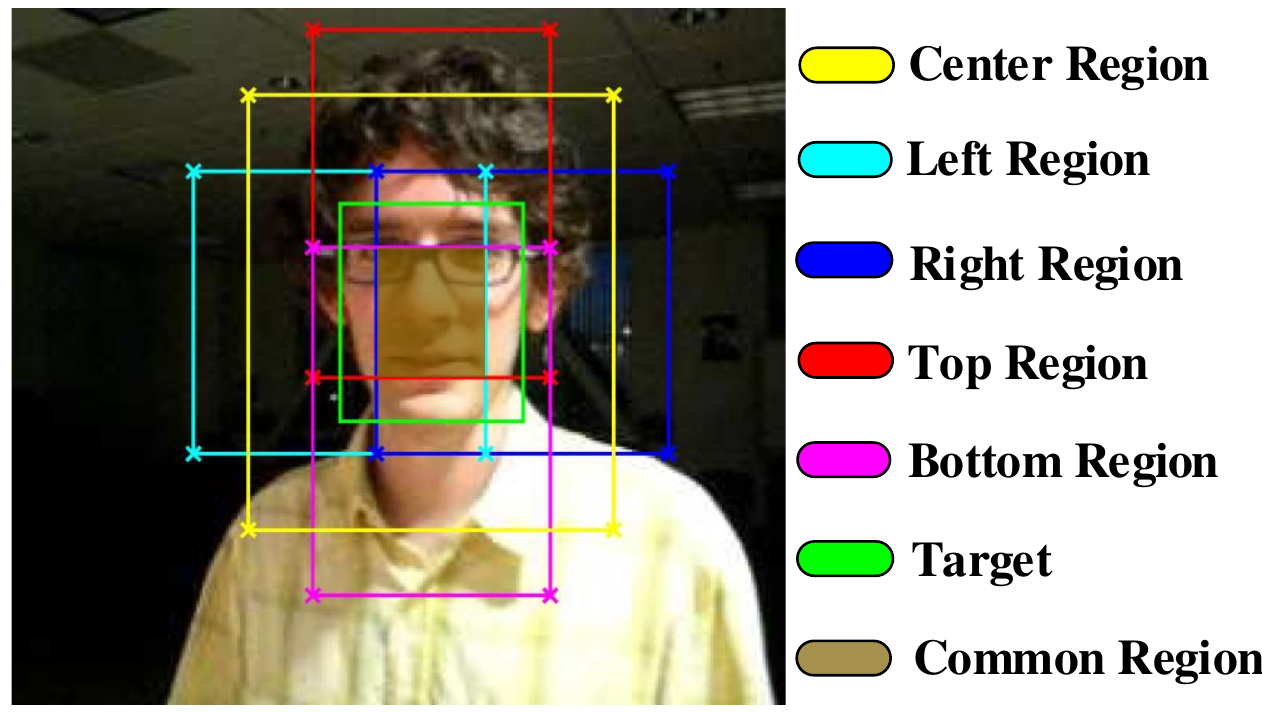}}\
\caption{An illustration of the generated center, boundary and common region based on the target bounding box.}
\label{fig:CommmonRegion}
\end{figure}
\begin{figure}[htbp]
\centering
\setlength{\abovecaptionskip}{0.1cm}
\setlength{\belowcaptionskip}{-0.4cm}
\subfloat[]{%
  \includegraphics[width=0.445\linewidth]{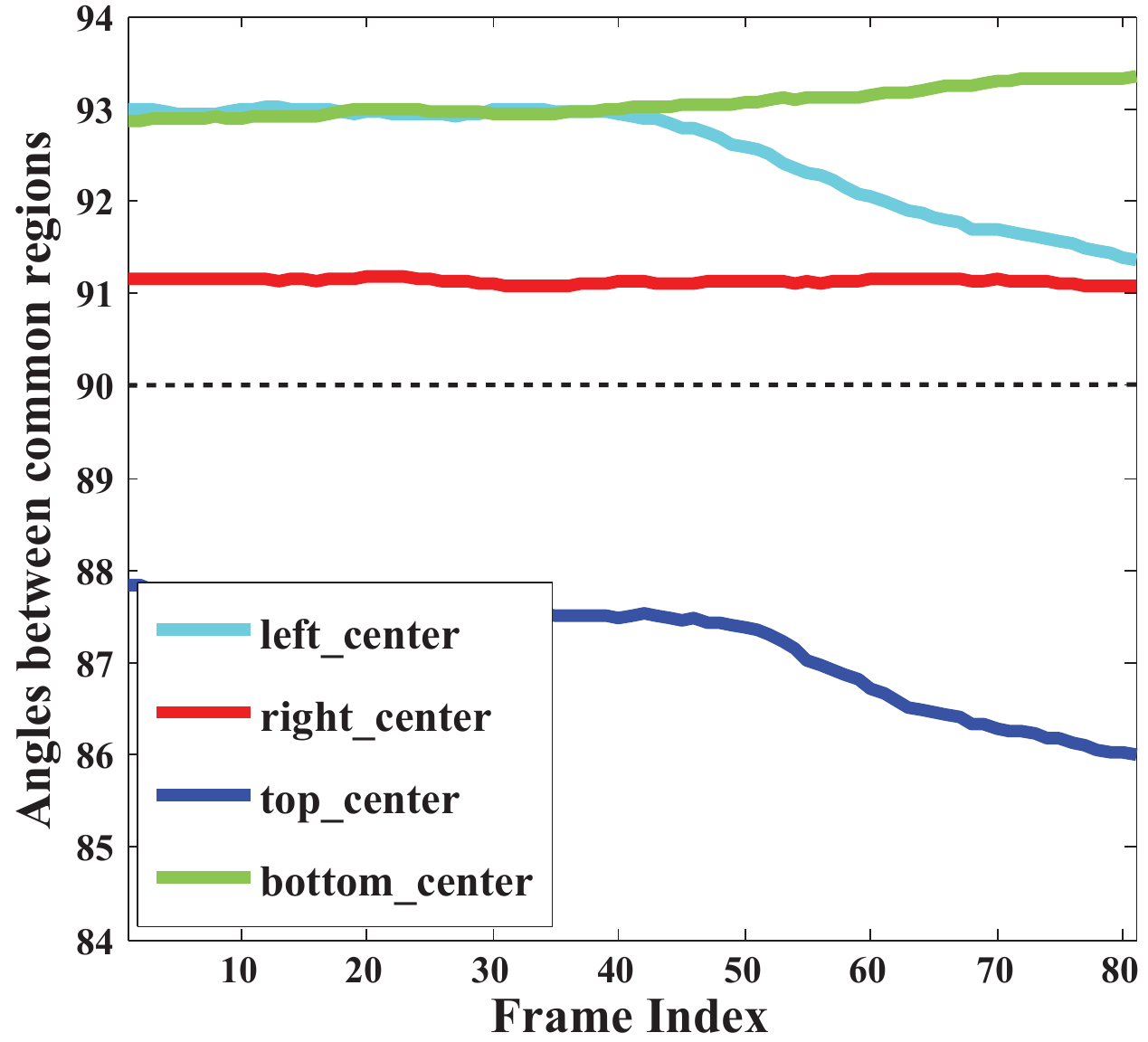}}\
\subfloat[]{%
  \includegraphics[width=0.45\linewidth]{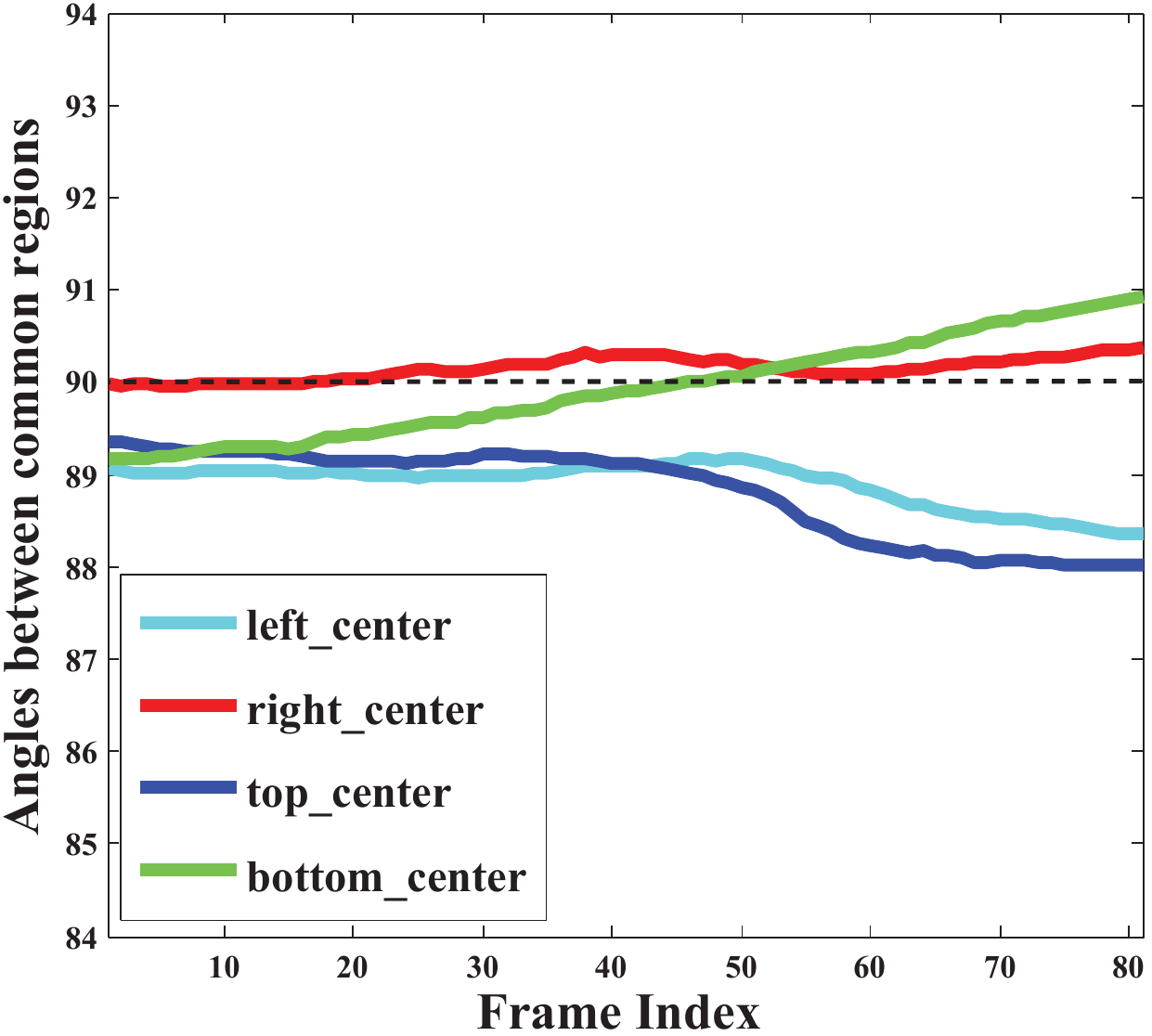}}\
\caption{The angles (in degree) between the common regions of center and boundary CFs on the sequence \emph{Skiing}. (a) The angles computed by training CCF and BCFs independently.
(b) Using the near-orthogonality constraint, the angles obtained by IBCCF can be more near to $90^\circ$, and 4.7\% gains by overlap precision is attained during tracking.}
\label{fig:AngleResults}
\end{figure}
On the other hand, the filtering responses of center CF achieve its maximum at the center position, and thus the angle between $[\mathbf{x}]_{\text{vec}}$ and $[\mathbf{w}]_{\text{vec}}$ should be small.
Therefore, $[\mathbf{w}_l]_{\text{vec}}$ should be nearly orthogonal with $[\mathbf{w}]_{\text{vec}}$.

However, in general the sizes of left boundary region and center region are not the same.
From Fig. \ref{fig:CommmonRegion}, one can see that they share a common region.
Let $[\tilde{\mathbf{w}}]_{\text{vec}}$ be the vectorization of center CF in the common region, and so do $[\tilde{\mathbf{x}}]_{\text{vec}}$ and $[\tilde{\mathbf{w}}_l]_{\text{vec}}$.
We then extend the near-orthogonality property to the common region, and expect that $[\tilde{\mathbf{w}}_l]_{\text{vec}}$ and $[\tilde{\mathbf{w}}]_{\text{vec}}$ are also nearly orthogonal, \ie, $[\tilde{\mathbf{w}}_l]_{\text{vec}}^T [\tilde{\mathbf{w}}]_{\text{vec}} \approx 0$.
Analogously, we also expect that $[\tilde{\mathbf{w}}_r]_{\text{vec}}^T [\tilde{\mathbf{w}}]_{\text{vec}} \approx 0$, $[\tilde{\mathbf{w}}_t]_{\text{vec}}^T [\tilde{\mathbf{w}}]_{\text{vec}} \approx 0$, $[\tilde{\mathbf{w}}_b]_{\text{vec}}^T [\tilde{\mathbf{w}}]_{\text{vec}} \approx 0$.
Fig. \ref{fig:AngleResults} shows the angles between the center CF and boundary CFs in common region on the sequence \emph{Skiing}. From Fig. \ref{fig:AngleResults}(a), one can note that it is roughly hold true that the boundary and center CFs are nearly orthogonal.
Thus, as illustrated in Fig. \ref{fig:AngleResults}(b), we expect better near-orthogonality and tracking accuracy can be attained by imposing the near-orthogonality constraint on the training of CCF and BCFs.
Empirically, for \emph{Skiing}, the introduction of near-orthogonality constraint does bring 4.7\% gains by overlap precision during tracking.

\subsection{Problem Formulation of IBCCF}
By enforcing the near-orthogonality constraint, we propose our IBCCF model to integrate boundary and center CFs, resulting in the following objective,
\begin{eqnarray}\label{equ:IBCCF}
\arg\min\limits_{\mathbf{W}} \sum_{k \in \Psi} \mathcal{L}_k(\mathbf{w}_{k}) + \mathcal{E}(\mathbf{w}) + \mu \sum_{k \in \Psi} \left ( [\tilde{\mathbf{w}}]_{\text{vec}}^T [\tilde{\mathbf{w}}_k]_{\text{vec}}  \right )^2
\end{eqnarray}
where $\Psi = \{ l, r, t, b \}$, and $\mathbf{W} = \{ \mathbf{w}, \mathbf{w}_l, \mathbf{w}_r, \mathbf{w}_t, \mathbf{w}_b \}$.
$\mathcal{L}_k(\mathbf{w}_{k})$ and $\mathcal{E}(\mathbf{w})$ are defined in Eqns. (\ref{equ:BCFFomulation}) and (\ref{equ:CF}).

\textbf{Comparison with DSST \cite{danelljan2016discriminative}.} Even DSST also adopts 1D and 2D CFs, it learns separate 2D and 1D CFs for translation and scale estimation on scale pyramid representation, respectively.
And our IBCCF is distinctly different with DSST from three aspects:
(i) While DSST formulates scale estimation as 1D CF, BCF is among the first to suggest a novel parameterization of bounding box, and formulates boundary localization as 1D CFs.
(ii) For DSST, the inputs to 1D CF are image patches at different scales, while the input for BCF is four regions covering the edges of bounding boxes.
(iii) In DSST, the 1D CF and 2D translation CF are separately trained. While in IBCCF, 1D BCFs and 2D CCF are jointly learned by solving the IBCCF model in Eqn. (\ref{equ:IBCCF}).

\section{Optimization} \label{OPTIMATION}
In this section, we propose an ADMM method to minimize Eqn. (\ref{equ:IBCCF}) by alternately updating center CF $\mathbf{w}$ and
boundary CFs $\mathbf{w}_k$ ($k \in \Psi$), where each subproblem can be easily solved with a close-form solution.

We first employ variable splitting method to change Eqn. (\ref{equ:IBCCF}) into a linear equality constrained optimization problem:
\begin{eqnarray}\label{equ:IBCCFVS}
&\arg\min\limits_{\mathbf{W}} \sum\limits_{k \in \Psi} \mathcal{L}_k(\mathbf{w}_{k}) + \mathcal{E}(\mathbf{w}) + \mu \sum\limits_{k \in \Psi} \left ( [\tilde{\mathbf{w}}]_{\text{vec}}^T [\tilde{\mathbf{w}}_k]_{\text{vec}}  \right )^2 \\\nonumber
&\text{s.t.} \quad \mathbf{g} = \mathbf{w}, \mathbf{u}_k = \mathbf{w}_k, k \in \Psi.
\end{eqnarray}

Hence, the Augmented Lagrangian Method (ALM) can be applied to solve Eqn. (\ref{equ:IBCCFVS}), and its augmented lagrangian form \cite{boyd2011distributed} is reformulated as:
\begin{eqnarray}\label{equ:ALM}
\arg\min\limits_{\mathbf{H}} \sum_{k \in \Psi} \mathcal{L}_k(\mathbf{w}_{k}) + \mathcal{E}(\mathbf{w}) + \mu \sum_{k \in \Psi} \left ( [\tilde{\mathbf{g}}]_{\text{vec}}^T [\tilde{\mathbf{u}}_k]_{\text{vec}}  \right )^2\\\nonumber
+\rho \left \| \mathbf{w}-\mathbf{g}-\mathbf{p}\right \|^2 + \sum_{k \in \Psi} \gamma_k \left \| \mathbf{w}_k -\mathbf{u}_k -\mathbf{q}_k \right \|^2,
\end{eqnarray}
where $\mathbf{H} = \{\mathbf{W}, \mathbf{g}, \mathbf{p}, \mathbf{u}_{k}, \mathbf{q}_{k}\}$, $\mathbf{p}$ and $\mathbf{q}_k$ represent the Lagrange multiplier, $\rho$ and $\gamma_k$ are penalty factors, respectively.
For multivariable non-convex optimization, ADMM iteratively updates one of variables while keeping the rest fixed, hence the convergence can be guaranteed \cite{boyd2011distributed}.
By using ADMM, Eqn. (\ref{equ:ALM}) is divided into the following subproblems:
\begin{align}\label{equ:ADMM}
\begin{cases}
   \mathbf{w}^{(i+1)}&=\arg\min\limits_{\mathbf{w}} \mathcal{E}(\mathbf{w})+ \rho^{(i)}\left \| \mathbf{w} - \mathbf{g}^{(i)} - \mathbf{p}^{(i)}\right \|^2 \\
   \mathbf{g}^{(i+1)}&=\arg\min\limits_\mathbf{g} \mu {\left\|(\mathbf{S}^{(i)})^T[\tilde{\mathbf{g}}]_{\text{vec}}\right\|^2}\\
                         &\quad+\rho^{(i)}{\left\| {\mathbf{w}^{\left ( i \right )} - \mathbf{g} - \mathbf{p}^{\left ( i \right )}} \right\|^2}\\
   \mathbf{w}_k^{(i+1)}&=\arg\min\limits_{\mathbf{w}_k} \mathcal{L}(\mathbf{w}_k)+ \gamma_k^{(i)}\left \| \mathbf{w}_k - \mathbf{u}_k^{(i)} - \mathbf{q}_k^{(i)}\right \|^2 \\
   \mathbf{u}_k^{(i+1)}&=\arg\min\limits_{\mathbf{u}_k} \mu {\left\|[\tilde{\mathbf{g}}^{(i)}]_{\text{vec}}^T [\tilde{\mathbf{u}}_k]_{\text{vec}}\right\|^2}\\
                          &\quad+\gamma_k^{(i)}{\left\| {\mathbf{w}_k^{\left ( i \right )} - \mathbf{u}_k - \mathbf{q}_k^{\left ( i \right )}} \right\|^2}\\
   \mathbf{p}^{\left (i+1 \right )}&=\mathbf{p}^{\left ( i \right )} + \mathbf{g}^{\left ( i+1 \right )} - \mathbf{w}^{\left ( i+1 \right )}\\
   \mathbf{q}_k^{\left (i+1 \right )}&=\mathbf{q}_k^{\left ( i \right )} + \mathbf{u}_k^{\left ( i+1 \right )} - \mathbf{w}_k^{\left ( i+1 \right )}\\
\end{cases}
\end{align}
where $\mathbf{S}^{(i)}=[[\tilde{\mathbf{u}}_l^{(i)}]_{\text{vec}}, [\tilde{\mathbf{u}}_r^{(i)}]_{\text{vec}}, [\tilde{\mathbf{u}}_t^{(i)}]_{\text{vec}}, [\tilde{\mathbf{u}}_b^{(i)}]_{\text{vec}}]$ and $k\in\Psi$.
From Eqn. (\ref{equ:IBCCF}), we can see that the boundary CFs are independent of each other, so each pair of $\mathbf{w}_k$ and $\mathbf{u}_k$ can be updated in parallel for efficiency.
Next, we detail the solution to each subproblem as follows:\\
\textbf{Subproblem $\mathbf{w}$.} Using the properties of circulant matrix and FFT, the closed form solution of $\mathbf{w}$ is given as:\\
\begin{eqnarray}\label{equ:subw}
\mathbf{w}= \mathcal{F}^{-1}\left (\frac{\hat{\mathbf{x}}^{*}\odot \hat{\mathbf{y}} +\hat{\mathbf{g}}+\hat{\mathbf{p}}}{\hat{\mathbf{x}}^{*} \odot \hat{\mathbf{x}} +\lambda+\rho} \right )
\end{eqnarray}
\textbf{Subproblem $\mathbf{g}$.} The second row of Eqn. (\ref{equ:ADMM}) is rewritten as:
\begin{eqnarray}\label{equ:subg}
\arg\min_{[\mathbf{g}]_{\text{vec}}} \mu{\left\| \mathbf{Q}^T[\mathbf{g}]_{\text{vec}} \right\|^2} + \rho {\left\| [\mathbf{w}]_{\text{vec}} - [\mathbf{g}]_{\text{vec}} - [\mathbf{p}]_{\text{vec}} \right\|^2}
\end{eqnarray}
where the matrix $\mathbf{Q}=\begin{bmatrix}\mathbf{S}\\\mathbf{0}\end{bmatrix} $ is obtained by padding zeros to each column of $\mathbf{S}$. Then $[\mathbf{g}]_{\text{vec}}$ can be computed by:
\begin{eqnarray}\label{equ:solutionvg}
[\mathbf{g}]_{\text{vec}} = \left( \mu \mathbf{Q}\mathbf{Q}^T + \rho I \right)^{-1}\rho \left( [\mathbf{w}]_{\text{vec}} - [\mathbf{p}]_{\text{vec}} \right)
\end{eqnarray}

Note that matrix $\mathbf{Q}$ only contains four columns, thus Singular Value Decomposition (SVD) can be used for improving the efficiency. By performing SVD of $\mathbf{Q}$ with $\mathbf{Q}= \mathbf{U}\Sigma \mathbf{V}^T$, we have:
\begin{eqnarray}\label{equ:solutionvg1}
[\mathbf{g}]_{\text{vec}} = \mathbf{U}\Lambda {\mathbf{U}^T}\left([\mathbf{w}]_{\text{vec}} - [\mathbf{p}]_{\text{vec}}\right)
\end{eqnarray}
where $\Lambda=(\frac{\mu}{\rho}\Sigma\Sigma^T + \mathbf{I})^{-1}$. Let the nonzero elements in matrix $\Sigma$ be $[\lambda_1,\lambda_2,\lambda_3,\lambda_4]$,
the nonzero elements of diagonal matrix $\Lambda$ become $[\frac{\rho}{\mu \lambda_1^2 + \rho},\frac{\rho}{\mu \lambda_2^2 + \rho}, \frac{\rho}{\mu \lambda_3^2 + \rho}, \frac{\rho}{\mu \lambda_4^2 + \rho}, ..., 1]$.
Hence, Eqn. (\ref{equ:solutionvg1}) can be written as:
\begin{eqnarray}\label{equ:solutionvg2}
[\mathbf{g}]_{\text{vec}} = \left(\mathbf{I} - \mathbf{U}(\mathbf{I} - \Lambda )\mathbf{U}^T \right)\left([\mathbf{w}]_{\text{vec}} - [\mathbf{p}]_{\text{vec}}\right)
\end{eqnarray}

\begin{figure}[htb]
\centering
\setlength{\abovecaptionskip}{0.1cm}
\setlength{\belowcaptionskip}{-0.5cm}
\subfloat{%
  \includegraphics[width=0.22\textwidth]{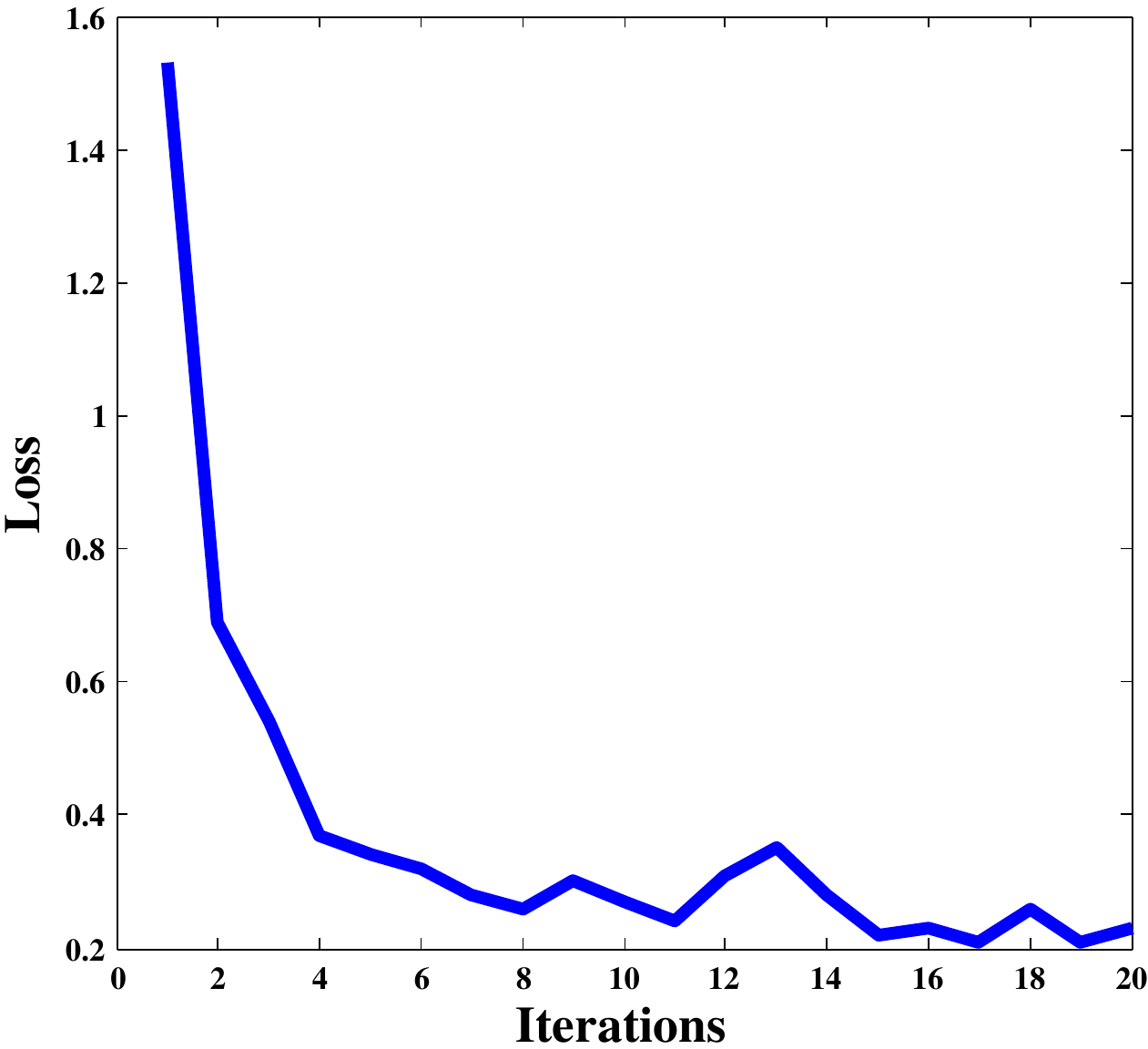}}\
\caption{Convergence curve of the proposed IBCCF with ADMM on the fifth frame of sequence \emph{Skiing} on OTB benchmark.}
\label{fig:losscurve}
\end{figure}

Since diagonal matrix $(\mathbf{I}-\Lambda)$ only contains four nonzero elements, we have $\mathbf{U}(\mathbf{I}-\Lambda)\mathbf{U}^T = \hat{\mathbf{U}}\text{Diag}(f)\hat{\mathbf{U}}^T$, where $\hat{\mathbf{U}}$ is the first four columns of matrix $\mathbf{U}$
and $\text{Diag}(f)$ denotes the diagonal matrix of the nonzero elements in $(\mathbf{I}- \Lambda)$. Such special case can be solved efficiently \footnote{Please refer to SVD function with ``economy'' mode in Matlab.}.\\
\textbf{Subproblem $\mathbf{w}_k$.} The solution of $\mathbf{w}_k$ shares similar solution with one of Eqn. (\ref{equ:subw}):\\
\begin{eqnarray}\label{equ:subwl}
\mathbf{w}_k= \mathcal{F}^{-1}\left (\frac{\hat{\mathbf{x}_k}^{*}\odot \hat{\mathbf{y}_k} +\hat{\mathbf{u}_k}+\hat{\mathbf{q}_k}}{\hat{\mathbf{x}_k}^{*} \odot \hat{\mathbf{x}_k} +\lambda+\gamma_k} \right )
\end{eqnarray}
\textbf{Subproblem $\mathbf{u}_k$.} The fourth row of Eqn. (\ref{equ:ADMM}) is written as:
\begin{equation}\label{equ:subh}
\begin{aligned}
{[\mathbf{u}_k]}_{\text{vec}}=&\arg\min\limits_{[\mathbf{u}_k]_{\text{vec}}} \mu{\left\|[\mathbf{s}]_{\text{vec}}^T[\mathbf{u}_k]_{\text{vec}} \right\|^2} \\
&+\gamma_k {\left\| [\mathbf{w}_k]_{\text{vec}} - [\mathbf{u}_k]_{\text{vec}} - [\mathbf{q}_k]_{\text{vec}} \right\|^2}
\end{aligned}
\end{equation}
where $[\mathbf{s}]_{\text{vec}}=\begin{bmatrix}[\tilde{\mathbf{g}}]_{\text{vec}}\\\mathbf{0}\end{bmatrix}$ and the close-form solution of $[\mathbf{u}_k]_{\text{vec}}$ is:
\begin{eqnarray}\label{equ:solvevh}
[\mathbf{u}_k]_{\text{vec}}= (\mu[\mathbf{s}]_{\text{vec}}[\mathbf{s}]_{\text{vec}}^T+\gamma_k \mathbf{I})^{-1}\gamma_k([\mathbf{w}_k]_{\text{vec}}-[\mathbf{q}_k]_{\text{vec}})
\end{eqnarray}
Since $[\mathbf{s}]_{\text{vec}}[\mathbf{s}]_{\text{vec}}^T$ is rank-1 matrix, Eqn. (\ref{equ:solvevh}) can be efficiently solved with Sherman-Morrsion formula \cite{Pedersen2008Cookbook} , so we have:
\begin{eqnarray}\label{equ:solvevh1}
[\mathbf{u}_k]_{\text{vec}}= ( 1- \frac{\mu[\mathbf{s}]_{\text{vec}}[\mathbf{s}]_{\text{vec}}^T}{\gamma_k + \mu[\mathbf{s}]_{\text{vec}}^T[\mathbf{s}]_{\text{vec}}})([\mathbf{w}_k]_{\text{vec}}- [\mathbf{q}_k]_{\text{vec}})
\end{eqnarray}
\textbf{Convergence.} To verify the effectiveness of the proposed ADMM, we illustrate the convergence curve with ADMM on sequence
\emph{Skiing}. As shown in Fig. \ref{fig:losscurve}, although IBCCF model is a non-convex problem, we can see that it converges within very few iterations (four iterations in this case).
This phenomenon is ubiquitous in our experiments, and most of the sequences converges within five iterations.

\section{Experiments} \label{EXPERIMENTS}
\begin{table*}
\setlength{\abovecaptionskip}{0.1cm}
\setlength{\belowcaptionskip}{-0.5cm}
\centering
\scalebox{0.54}{
\begin{tabular}{cccccccccccccccccccc}
\toprule
&C-COT \cite{Danelljan2016CCOT}& SINT+ \cite{tao2016siamese}&SKSCF \cite{zuo2016learning}&Scale-DLSSVM \cite{ningobject}&Staple \cite{bertinetto2015staple}&SRDCF \cite{danelljan2015learning}&DeepSRDCF \cite{danelljan2015convolutional}&RPT \cite{Li2015RPT}&MEEM \cite{zhang2014meem}&DSST \cite{danelljan2016discriminative}&SAMF \cite{li2014scale}&HCF \cite{ma2015hierarchical}&SCF \cite{Liu2016SCF}&sHCF&IBCCF\\
\midrule
OTB-2013 & {\color{blue}83.4} & {\color{green}81.3} & 80.9 & 73.2 & 74.9 & 78 & 79.2 & 71.4 & 69.1 & 67.7 & 68.9 & 73.5 & 79.7 & 73.4 & {\color{red}83.7}\\
OTB-2015 & {\color{red}82.7} &  -   & 67.4 & 65.2 & 71.3 & 72.7 & {\color{green}77.6} & 64 & 62.3 & 62.2 & 64.5 & 65.6 & - & 69.2 & {\color{blue}78.4}\\
\bottomrule
\end{tabular}}
\caption{\small{Mean OP metrics (in \%) of different trackers on OTB-2013 and OTB-2015. The best three results are shown in {\color{red} red}, {\color{blue} blue} and {\color{green} green} fonts, respectively.}}
\label{tab:OTB2015}
\end{table*}
In this section, we first compare IBCCF with state-of-the-art trackers on OTB dataset \cite{wu2015object}.
Then we validate the effects of each component on IBCCF, and analyze the time cost using OTB dataset.
Finally, we conduct comparative experiments on Temple-Color \cite{Liang2015Encoding} and VOT benchmarks.

Following the common settings in HCF \cite{ma2015hierarchical}, we implement IBCCF by using the outputs of layers \emph{conv}3-4, \emph{conv}4-4 and \emph{conv}5-4
of VGG-Net-19 \cite{simonyan2014very} for feature extraction. To combine the responses from different layers, we follow the HCF setting and assign the weights of three layers in the center CF to 0.02, 0.5 and 1, respectively.
For boundary CFs, we omit the layer of \emph{conv}3-4 and set the weights for layers of \emph{conv}4-4 and \emph{conv}5-4 both to 1. The regularization parameters $\lambda$ and $\mu$ are set to
$10^{-4}$ and 0.1, respectively.
{Note that we employ a subset of 40 sequences from Temple-Color dataset as the validation set to choose the above parameters. Detailed description about the subset and corresponding experiments are given in Section \ref{Temple}.}
Our approach is implemented with Matlab by using MatConvNet Library. The average running time is about 1.25fps on a PC equipping with a Intel Xeon(R) 3.3GHz CPU, 32GB RAM and NVIDIA GTX 1080 GPU.
\subsection{OTB benchmark}
OTB benchmark consists of two subsets, i.e., OTB-2013 and OTB-2015.
OTB-2013 contains 51 sequences annotated with 11 different attributes, such as scale variation, occlusion and low resolution. OTB-2015 extends OTB-2013 to 100 videos.
We quantitatively evaluate our method with One-Pass Evaluation (OPE) protocol, where overlap precision (OP) metrics is used by computing the fraction of frames with bounding box overlaps exceeding 0.5 in a sequence.
Besides, we also provide overlap success plots containing the OP metrics over a range of thresholds.
\subsubsection{Comparison with state-of-the art trackers}
We compare our algorithm with 13 state-of-the-art methods: HCF \cite{ma2015hierarchical}, C-COT \cite{Danelljan2016CCOT}, SRDCF \cite{danelljan2015learning}, DeepSRDCF \cite{danelljan2015convolutional}, SINT+ \cite{tao2016siamese}, RPT \cite{Li2015RPT}, SCF \cite{Liu2016SCF},
SAMF \cite{li2014scale}, Scale-DLSSVM \cite{ningobject}, Staple \cite{bertinetto2015staple}, DSST \cite{danelljan2016discriminative}, MEEM \cite{zhang2014meem} and SKSCF \cite{zuo2016learning}.
Among them, most trackers except HCF and MEEM perform scale estimation during tracking. And both RPT and SCF methods exploit part-based models.
In addition, for verifying the effectiveness of BCFs on handling aspect ratio variation, we implement a HCF variant with five scales (denoted by sHCF) under the SAMF framework.
Note that we employ publicly available codes of compared trackers or copy the results from the original paper for fair comparison.

Table \ref{tab:OTB2015} lists a comparison of mean OP on OTB-2013 and OTB-2015 datasets. From it we can draw the following conclusions:
{(1) Our IBCCF outperforms most trackers except C-COT \cite{Danelljan2016CCOT} and surpasses its counterpart HCF (i.e., only center CF) by 10.2\% and 12.8\% on OTB-2013 and OTB-2015, respectively. We owe these significant improvements to integration of boundary CFs.
C-COT achieves higher mean OP than IBCCF on OTB-2015. It should be noted that spatial regularization is considered to suppress the boundary effect in both DeepSRDCF \cite{danelljan2015convolutional} and C-COT. Furthermore, C-COT also extends DeepSRDCF by learning multiple convolutional filters in continuous spatial domain.
%
%
In contrast, our IBCCF does not consider the spatial regularization and continuous convolution, and can yield favorable performance against the competing trackers.}
(2) our IBCCF is consistently superior to sHCF and other scale estimation based methods (e.g., DeepSRDCF) on both datasets. It indicates our boundary CFs are more helpful than scale estimation to CF-based trackers.
%
%
(3) Compared with part-based trackers (e.g., SCF), our IBCCF also shows its superiority, i.e., 4\% gains over SCF on OTB-2013 dataset.

\begin{figure}[htb]
\setlength{\abovecaptionskip}{0.2cm}
\setlength{\belowcaptionskip}{-0.4cm}
\centering
\subfloat{%
  \includegraphics[width=0.235\textwidth]{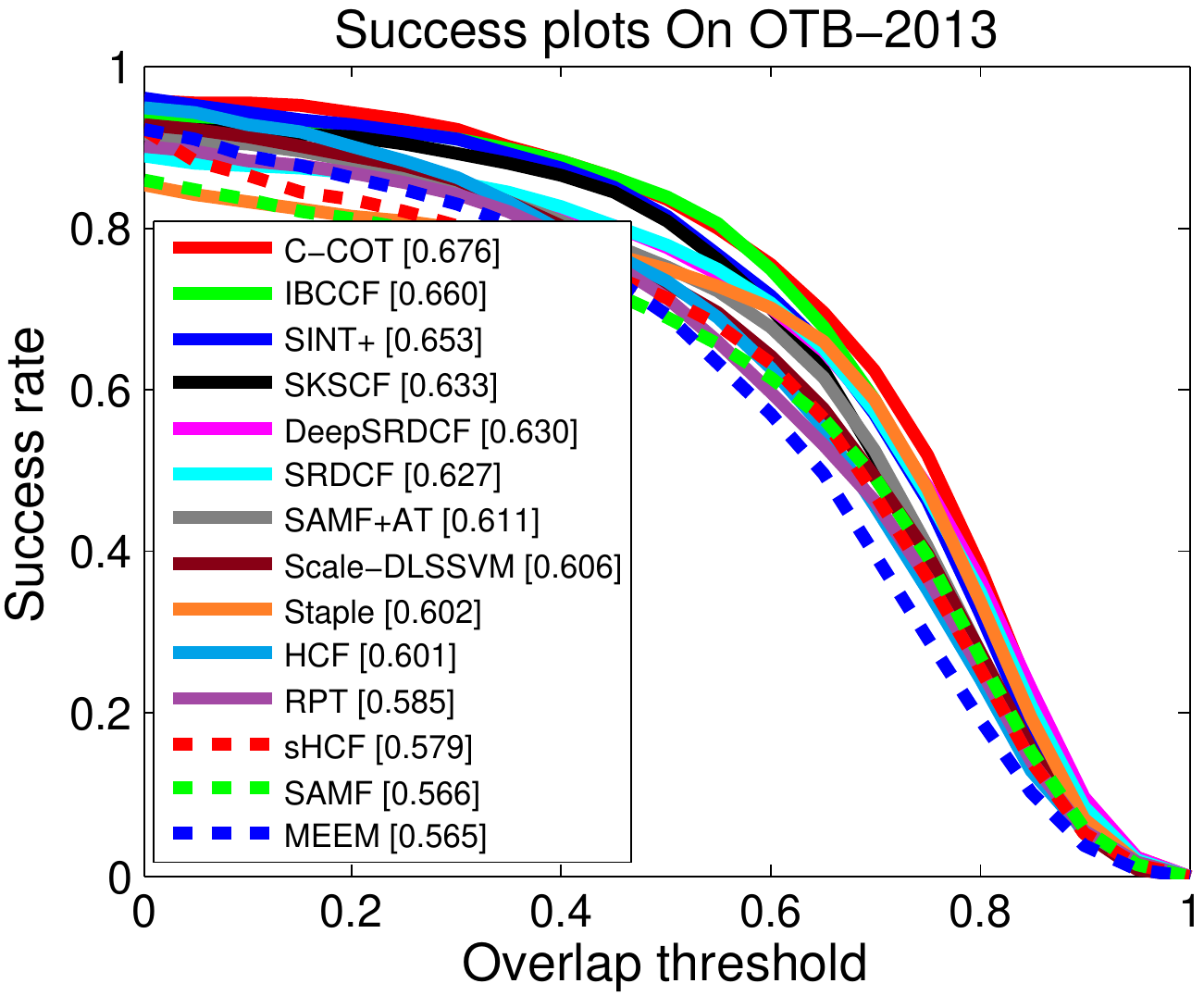}}\
\subfloat{%
  \includegraphics[width=0.233\textwidth]{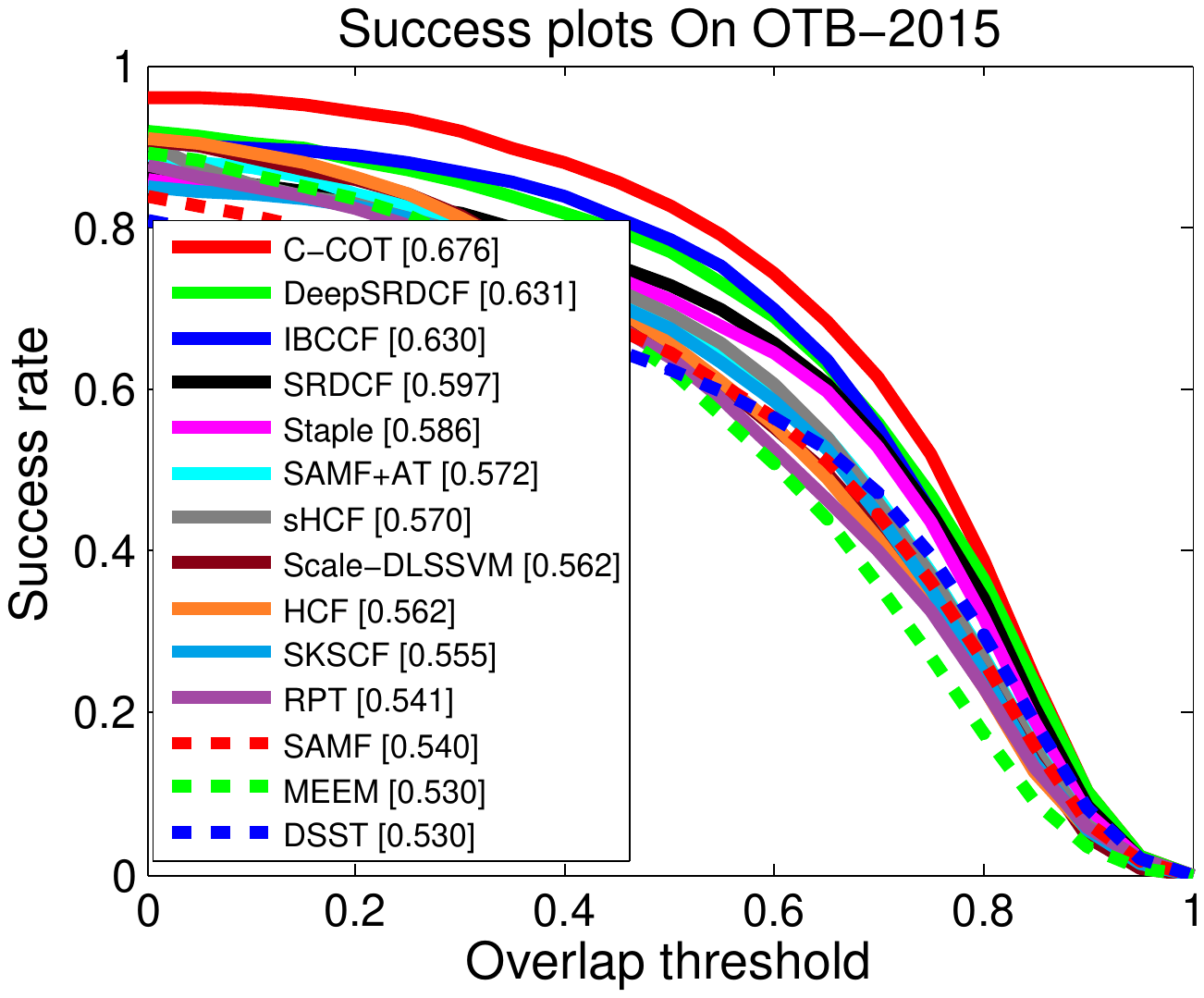}}\\
\caption{Comparison of overlap success plots with state-of-the-art trackers on OTB-2013 and OTB-2015.}
\label{fig:OPfig}
\end{figure}

Next, we show the overlap success plots of different trackers, which are ranked using the Area-Under-the-Curve (AUC) score.
As shown in Fig. \ref{fig:OPfig}, our IBCCF tracker is among the top three trackers on both datasets and outperforms HCF by 5.9\% and 6.8\% on OTB-2013 and OTB-2015 datasets, respectively.
%
\begin{figure}[htb]
\setlength{\abovecaptionskip}{0.2cm}
\setlength{\belowcaptionskip}{-0.3cm}
\centering
\subfloat{%
  \includegraphics[width=0.235\textwidth]{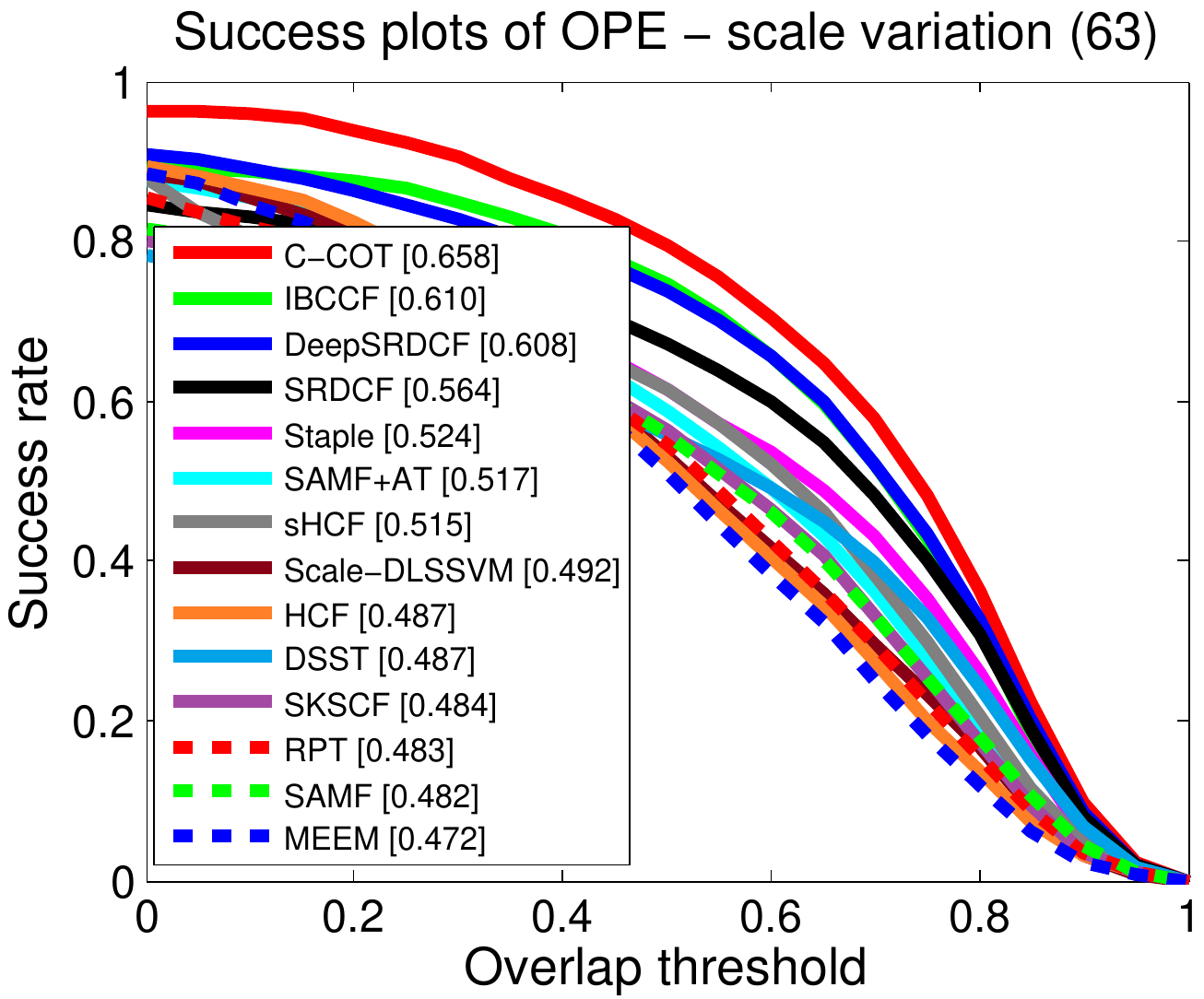}}\
\subfloat{%
  \includegraphics[width=0.235\textwidth]{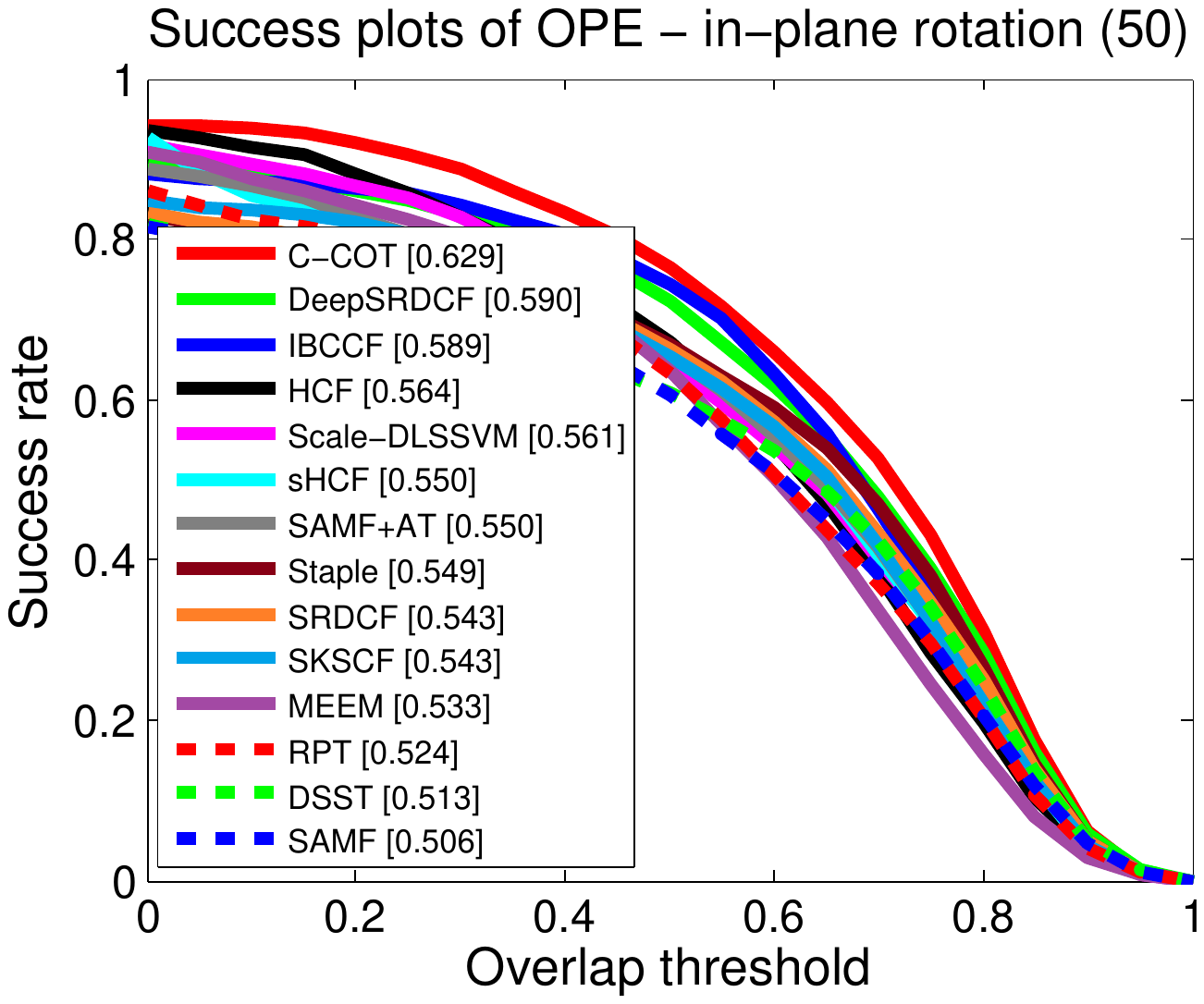}}\\[1mm]
\subfloat{%
  \includegraphics[width=0.237\textwidth]{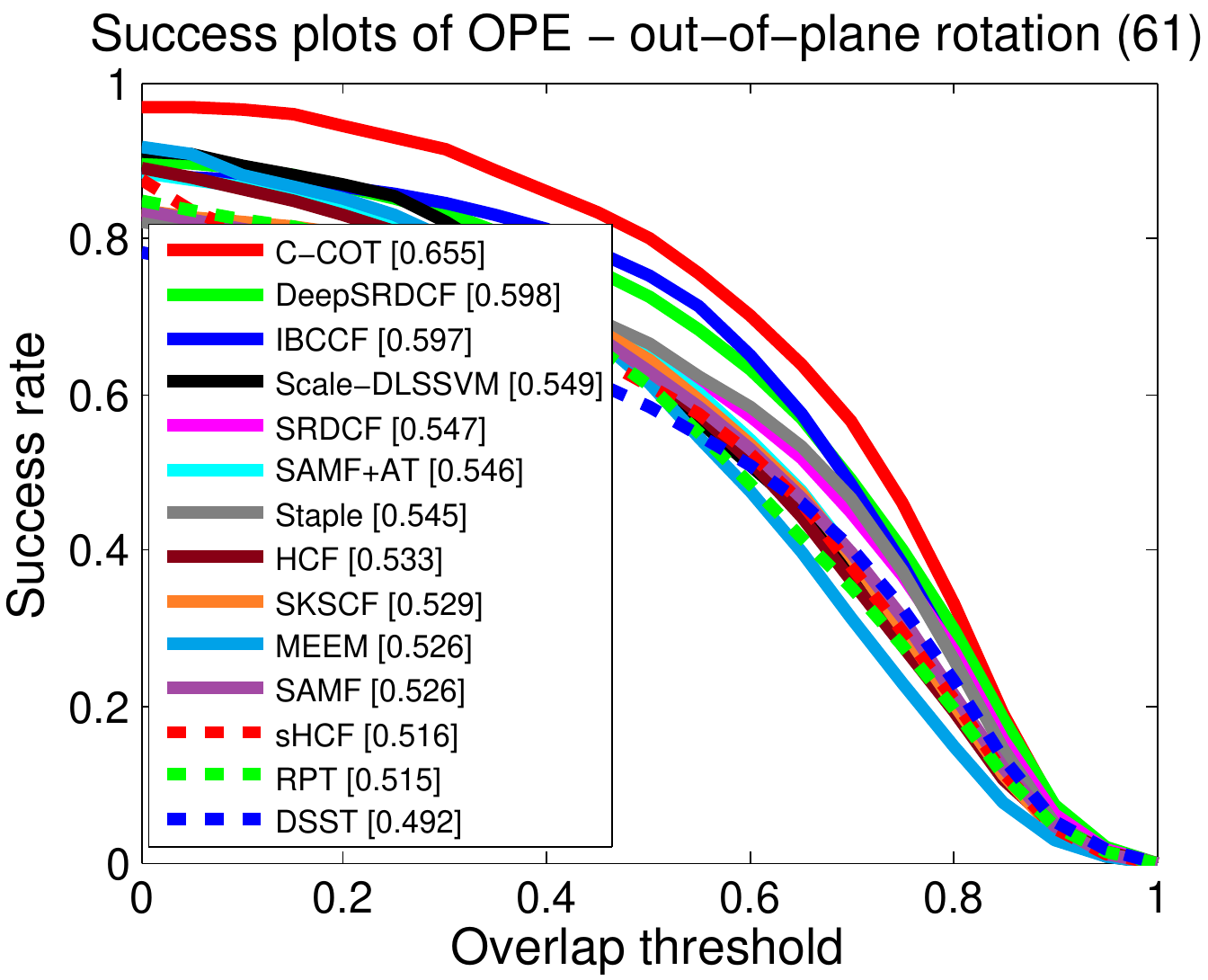}}\
\subfloat{%
  \includegraphics[width=0.235\textwidth]{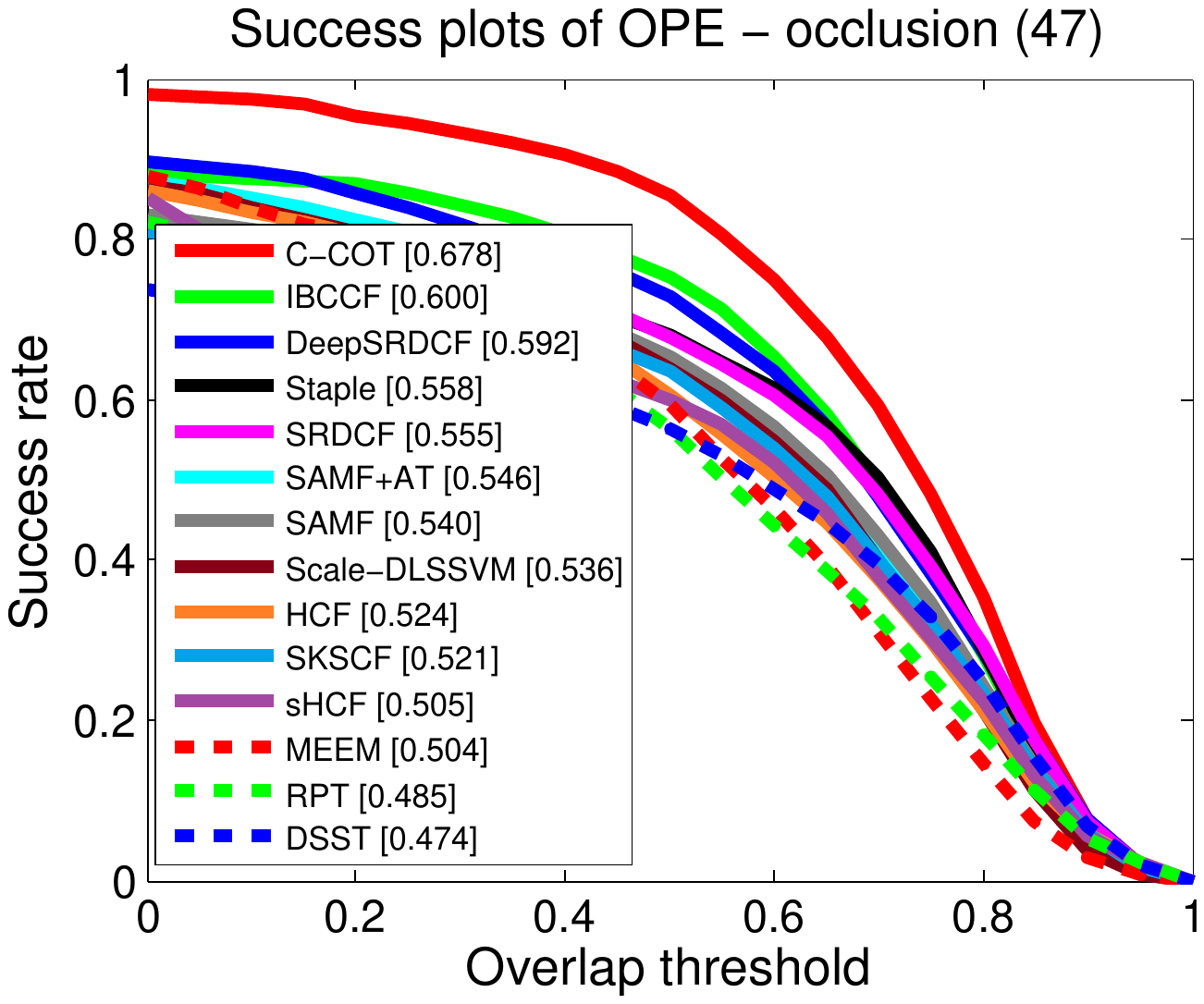}}\\
\caption{Overlap success plots of all competing trackers with four attributes great influencing aspect ratio variation on the OTB-2015.}
\label{fig:Attributes}
\end{figure}
\subsubsection{Video attributes related to aspect ratio variation}
In this subsection, we perform analysis of attributes influencing aspect ratio variation on OTB-2015 dataset.
Here we only provide the overlap success plots for four attributes which have great influence on aspect ratio variation and the rest results can be found in the supplementary material.

\textbf{\emph{Scale Variations}}: In the case of scale variations, target size continuously changes during tracking. It is worth noting that most of videos in this attribute only undergo scale changes rather than aspect ratio variation.
Despite in such setting, as illustrated in the upper left of Fig. \ref{fig:Attributes}, IBCCF still performs favorably among the compared trackers and is superior to the sHCF method by 9.5\%, demonstrating its effectiveness on handling target size variations.

\textbf{\emph{In-plane/Out-of-plane Rotation}}: In this case, targets encounter with rotation due to fast motion or viewpoint changes, which often cause aspect ratio variation of targets.
As shown in the upper right and lower left of Fig. \ref{fig:Attributes}, our IBCCF is robust to such kinds of variations and outperforms most of other trackers.
Specifically, IBCCF achieves remarkable improvements over its counterparts HCF, i.e., 2.5\% and 6.4\% gains in case of in-plane and out-of-plane rotation, respectively.
It indicates our IBCCF can deal with aspect ratio variation caused by rotations.

\textbf{\emph{Occlusion}}: Obviously, the partial occlusions can lead to aspect ratio variation of target. And complete occlusions also have an adverse impact on the boundary prediction.
Despite of these negative effects, IBCCF still outperforms most of the competing trackers and brings 7.6\% gains over the center CF tracker (\ie HCF).
\subsection{Internal Analysis of the proposed approach}
\subsubsection{\leftline{Impacts of Boundary CFs and near-orthogonality}}
Here we investigate the impact of boundary CFs and near-orthogonality property on the proposed IBCCF approach. To achieve this, we make four different variants of IBCCF: the tracker only with 1D boundary CFs (BCFs), the tracker only with the center CF (i.e., HCF), the IBCCF  tracker without orthogonality constraints denoted by IBCCF (w/o constraint) and full IBCCF model.
Table \ref{tab:component} summarizes the  mean OP and AUC score of four methods on OTB-2015.

From Table \ref{tab:component}, one can see that both BCFs tracker and orthogonality constraint are key parts of the proposed IBCCF method, and they can bring significant improvements over the center CF tracker.
Detailed analysis on the results can be found in the supplementary material.
%
%
%
%

\begin{table}
\scriptsize
\setlength{\abovecaptionskip}{-0.1cm}
\setlength{\belowcaptionskip}{-0.4cm}
\begin{center}
\scalebox{1.2}{
\begin{tabular}{ccccccccccc}
\toprule
 &Mean OP&AUC Score\\
\midrule
BCFs &50.2&39\\
Center CF &65.6&56.2\\
IBCCF (w/o constraints) &72&58.9\\
IBCCF &78.4&63\\
\bottomrule
\end{tabular}}
\end{center}
\caption{\small{Evaluation results of component experiments on both mean OP and AUC score metrics (in \%).}}
\label{tab:component}
\end{table}

\subsubsection{Time Analysis}
Here we analyze the average time cost of IBCCF for each stage on OTB-2015 dataset. The results are shown in Table \ref{tab:Time}.
One can clearly see that 
all subproblems including $\mathbf{g}$ and $\mathbf{u_k}$ can be solved rapidly, validating the efficiency of the ADMM solution.
%
%
Overall, the average running time of IBCCF and IBCCF (w/o constraint) is about 1.25 and 2.19 fps on OTB-2015 dataset, respectively.
\begin{table}
\scriptsize
\setlength{\abovecaptionskip}{-0.2cm}
\setlength{\belowcaptionskip}{-0.4cm}
\begin{center}
\scalebox{1.2}{
\begin{tabular}{ccccccccccc}
\toprule
&Time Cost(ms)\\
\midrule
CCF Feature Extraction & 95\\
BCFs Feature Extraction &141\\
CCF Prediction &26\\
BCFs Prediction & 40\\
Subproblem $\mathbf{w}$ & 51\\
Subproblem $\mathbf{g}$ & 37 \\
Subproblem $\mathbf{w_k}$ & 40 \\
Subproblem $\mathbf{u_k}$ & 4 \\
\bottomrule
\end{tabular}}
\end{center}
\caption{\small{Time cost of IBCCF for each stage on OTB-2015 dataset. Note that all the time listed above is measured in milliseconds.}}
\label{tab:Time}
\end{table}
\begin{table*}[!htb]
\setlength{\abovecaptionskip}{0.2cm}
\setlength{\belowcaptionskip}{-0.5cm}
\scalebox{0.63}{
\centering
\begin{tabular}{ccccccccccccccccc}
\toprule
 & DSST  \cite{danelljan2016discriminative}& sKCF & Struck \cite{hare2011struck} &CCOT \cite{Danelljan2016CCOT}& SRDCF \cite{danelljan2015learning}& DeepSRDCF \cite{danelljan2015convolutional}& Staple \cite{bertinetto2015staple}& MDNet\_N \cite{nam2016mdnet} & TCNN \cite{NamBH16}& DPT& SMPR & SHCT& HCF \cite{ma2015hierarchical}& IBCCF \\
\midrule
EAO &0.181&0.153&0.142&0.331&0.247&0.276&0.295&0.257&0.325&0.236&0.147&0.266&0.220&0.266\\
\midrule
Accuracy &0.5&0.44&0.44&0.52&0.52&0.51&0.54&0.53&0.54&0.48&0.44&0.54&0.47&0.51\\
\midrule
Robustness & 2.72&2.75&1.5&0.85&1.5&1.17&1.35&1.2&0.96&1.75&2.78&1.42&1.38&1.22\\
\bottomrule
\end{tabular}}
\caption{\small{Comparison of different state-of-the-art trackers on VOT-2016 dataset.}}
\label{tab:VOT2015}
\end{table*}

\subsection{Temple-Color dataset} \label{Temple}
In this section, we perform comparative experiments on Temple-Color dataset which contains 128 color sequences.
%
Different from the OTB dataset, it contain more video sequences with aspect ratio changes.
%
%
Hence, to better exploit the potential of IBCCF, we also choose a subset of 40 sequences with the largest standard deviations of aspect ratio variation from Temple-Color dataset and compare IBCCF with other methods.
Note that the sequences in the subset are not overlapped with other datasets. In addition, for validating the effectiveness of IBCCF with hand-crafted features, we implement two variants of IBCCF with HOG and color name \cite{Danelljan2014Adaptive} features
(\ie IBCCF-HOGCN, IBCCF-HOGCN (w/o constraint)).

\begin{figure}[htb]
\setlength{\abovecaptionskip}{-0.0cm}
\setlength{\belowcaptionskip}{-0.4cm}
\centering
\subfloat[]{%
  \includegraphics[width=0.235\textwidth]{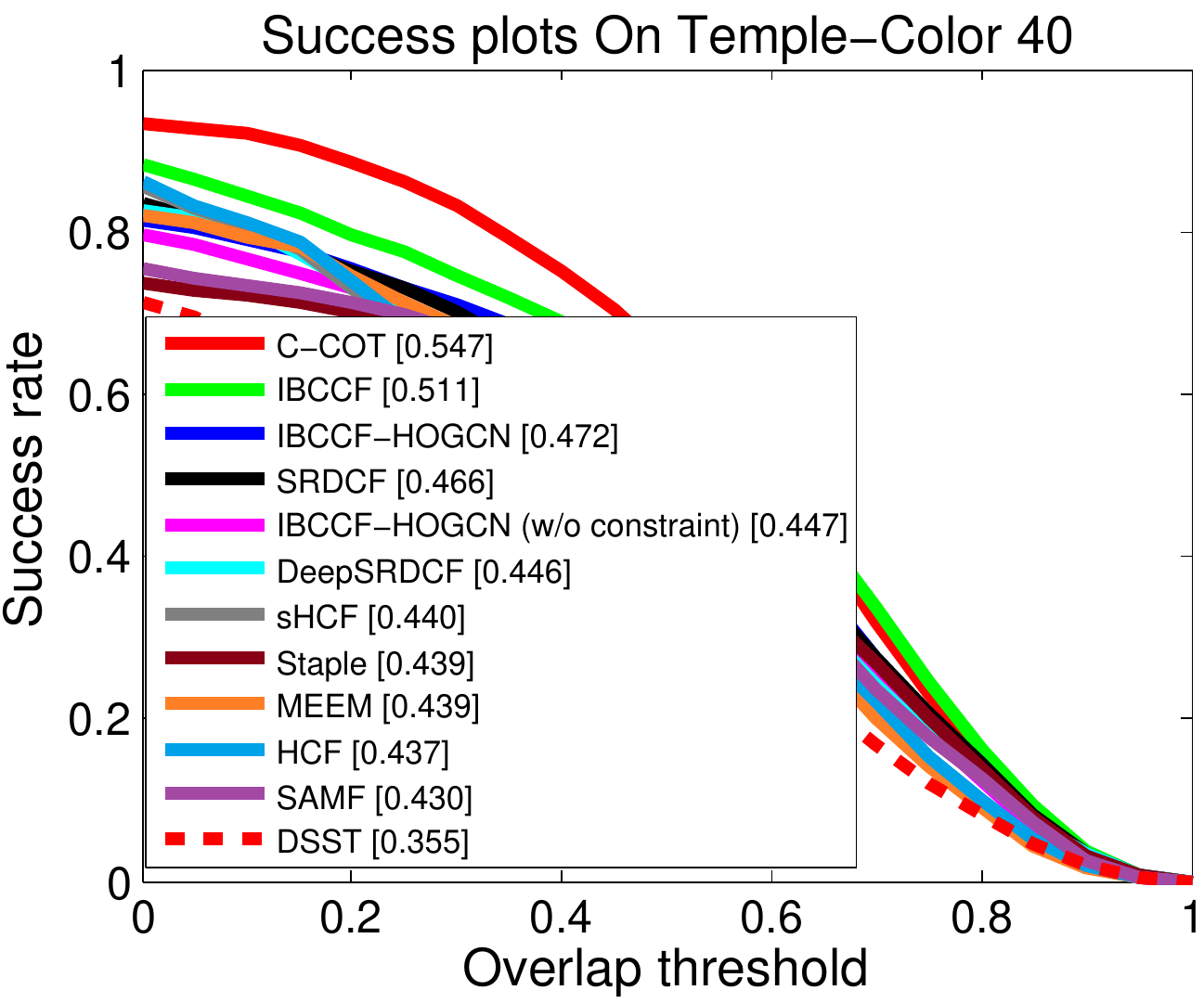}}\
\subfloat[]{%
  \includegraphics[width=0.235\textwidth]{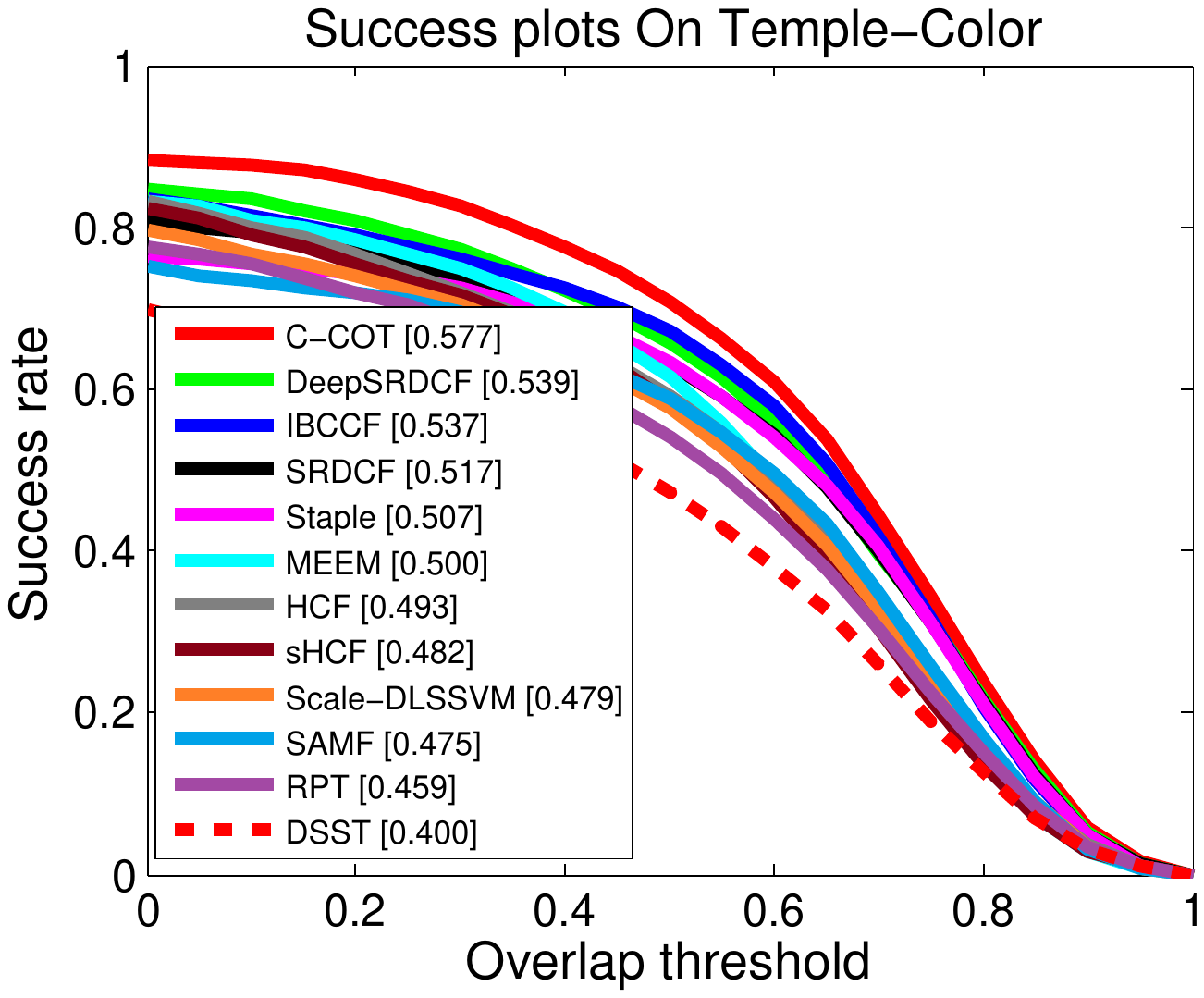}}\\
\caption{Comparison of the overlap success plots on two datasets. (a) the subset of 40 sequences from Temple-Color dataset. (b) the complete Temple-Color dataset.}
\label{fig:TempleColorResults}
\end{figure}

Fig. \ref{fig:TempleColorResults} illustrates the comparison of overlap success plots for different trackers on two datasets.
From Fig. \ref{fig:TempleColorResults}(a), one can see that IBCCF ranks the second among all trackers, demonstrating the effectiveness of IBCCF on handling aspect ratio variation again.
Furthermore, IBCCF-HOGCN also performs favorably against other methods and surpasses all of its counterparts (\ie IBCCF-HOGCN (w/o constraint), DSST and SAMF). This validates the superiority of IBCCF under hand-crafted feature setting.
%
%
As shown in Fig. \ref{fig:TempleColorResults}(b), IBCCF is among the top three best-performed trackers and outperforms its counterpart HCF by 4.4\%.
%
%
\subsection{The VOT benchmarks}
Finally, we conduct experiments on Visual Object Tracking (VOT) benchmark \cite{vcehovin2016visual}, which consist of 60 challenging videos from real-life datasets.
In VOT benchmark, a tracker is initialized at the first frame and reset again when it drifts the target. The performance is measured in terms of accuracy, robustness and expected average overlap (EAO). The accuracy computes the average overlap ratio between the estimated positions and ground
truth. The robustness score evaluates the average number of tracking failures. And the EAO metrics measures the average no-reset overlap of a tracker run on several short-term sequences.

\textbf{\emph{VOT-2016 results.}} We compare IBCCF with several state-of-the-art trackers, including MDNet \cite{nam2016mdnet} (VOT-2015 winner), TCNN \cite{NamBH16} (VOT-2016 winner)
and part-based trackers such as DPT, GGTv2, SMPR and SHCT. All the results are obtained from VOT-2016 challenge website\footnote{\url{http://www.votchallenge.net/vot2016/}}.
Table \ref{tab:VOT2015} lists the results on VOT-2016 dataset. One can note that IBCCF outperforms HCF method in terms of all three metrics. In addition, IBCCF also performs favorably
against the part-based trackers, validating the superiority of boundary tracking on handling aspect ratio variation.

\textbf{\emph{VOT-2017 results.}} At the time of writing, the results of VOT-2017 challenge were not available. Hence, we only report our results on the three metrics. In particular,
the EAO, accuracy and robustness scores of IBCCF on VOT-2017 dataset are 0.209, 0.48 and 1.57, respectively.

\section{Conclusion} \label{CONCLUSION}
In this work, we propose a tracking framework by integrating boundary and center correlation filters (IBCCF) to address the aspect ratio variation problem.
Besides tracking the target center, a family of 1D boundary CFs is introduced to localize the left, right, top and bottom boundaries, thus can adapt to the target scale and aspect ratio changes flexibly.
Furthermore, we analyze the near-orthogonality property between the center and boundary CFs, and impose an extra orthogonality constraint on the IBCCF model for improving the performance.
An ADMM algorithm is also developed to solve the proposed model.
We perform both qualitative and quantitative evaluation on four challenging benchmarks, and the results show that the proposed IBCCF approach perform favorably against several state-of-the-art trackers.
{Since we only employ the basic HCF model as the center CF tracker, in the future, we will incorporate with spatial regularization and continuous convolution to further improve our IBCCF.}

\section{Acknowledgements}
This work is supported by the National Natural Science Foundation of China (grant no.~61671182 and 61471082) and Hong Kong RGC General Research Fund (PolyU 152240/15E).
\
{\small
\bibliographystyle{ieee}
\bibliography{egbib}
}

\end{document}